\documentclass{article}
\usepackage{xcolor}
\usepackage[preprint]{corl_2026}

\usepackage{amsmath,amssymb}
\usepackage{graphicx}
\usepackage{booktabs}
\usepackage{multirow}
\usepackage{subcaption}
\usepackage{float}

\usepackage{cleveref}

\newcommand{\name}{FM-VLA}
\newcommand{\paravspace}{\vspace{-8pt}}

\title{\name{}: Force-based Memory for Vision-Language-\\Action Models in Contact-Rich Manipulation}

\author{%
    Ruicheng Li$^{1,2}$\thanks{Work done during internship at Microsoft Research} 
    \quad 
    Qixiu Li$^{1,2}$\footnotemark[1]
    \quad 
    Ruichun Ma$^{2}$
    \quad 
    Yu Deng$^{2}$ 
    \quad
    Lin Luo$^{2}$
    \quad
    \textbf{Zhiying Du}$^{3,2}$\footnotemark[1] 
    \\
    \textbf{Jianfeng Xiang}$^{1,2}$\footnotemark[1]
    \quad
    \textbf{Huizhi Liang}$^{1,2}$\footnotemark[1]
    \quad
    \textbf{Ruicheng Wang}$^{4,2}$\footnotemark[1]
    \quad
    \textbf{Jiaolong Yang}$^{2}$\thanks{Corresponding author} 
    \quad
    \textbf{Baining Guo}$^{2}$
    \\
	$^1${Tsinghua University}  \quad $^2${Microsoft Research} \quad $^3${Fudan University}  \quad $^4${USTC}
}

\begin{document}
\maketitle
\vspace{-3.5em}
\begin{figure}[H]
  \centering
  \setlength{\abovecaptionskip}{5pt}
  \includegraphics[width=0.92\linewidth]{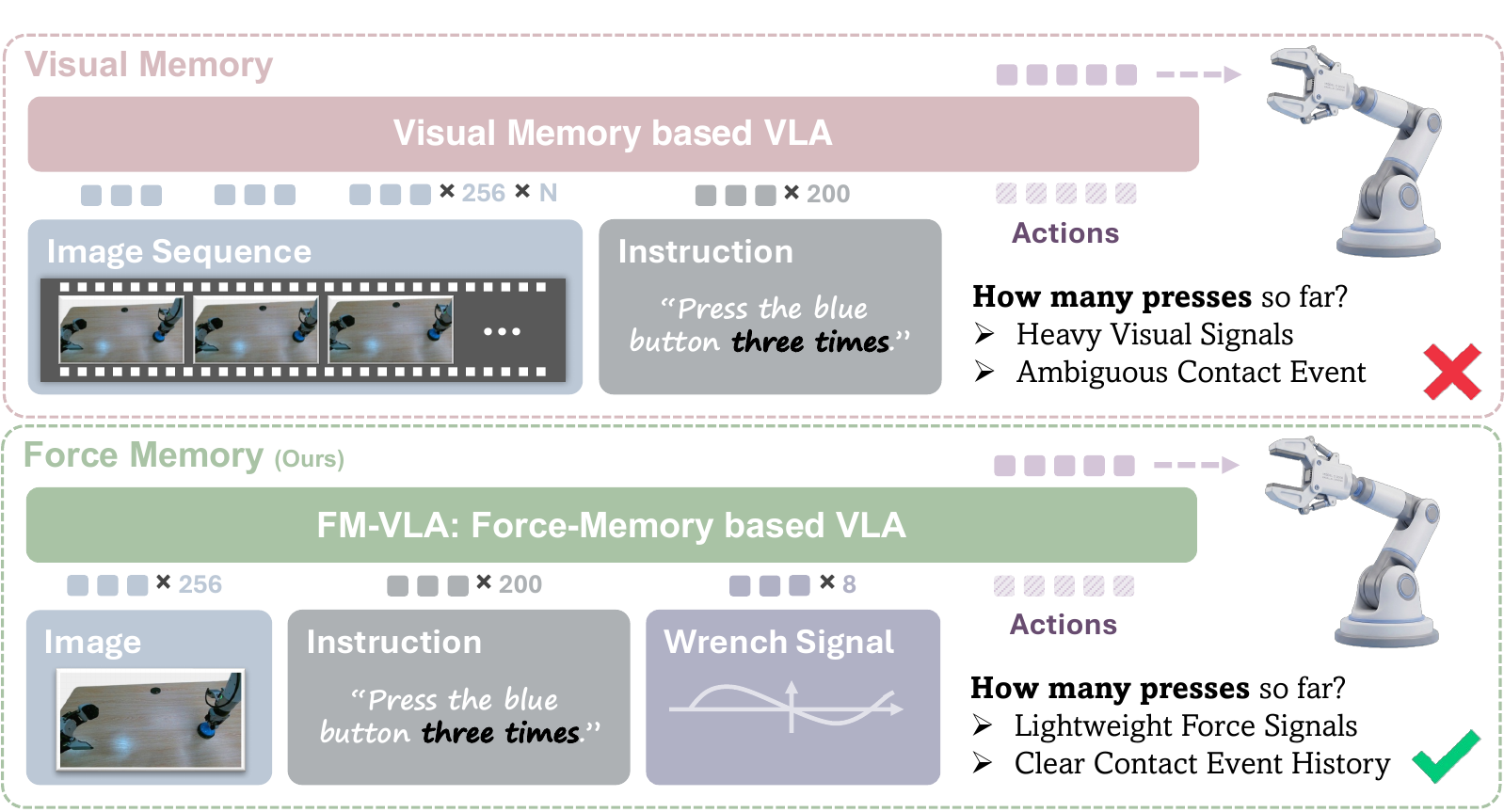}
  \caption{\small
  Comparison between visual memory based VLA and \name{}, which incorporates force (wrench) based memory to enable temporal context understanding for non-Markovian, contact-rich manipulation tasks.
  }
\label{fig:teaser}
\end{figure}

\begin{abstract}
Vision-language-action (VLA) models have achieved impressive generalization in robotic manipulation, and recent memory-augmented VLAs have relaxed the Markovian assumption by conditioning on past images or language summaries.
Vision-based memory approaches address this by conditioning on sampled past image frames, but they are computationally expensive and fundamentally limited when temporal events are visually ambiguous, e.g., pushing a button multiple times with small movements. 
We propose \name{}, a VLA model with force-based memory, enabling temporal context reasoning for non-Markovian, contact-rich manipulation.
We encode force histories into compact force memory tokens with a variational autoencoder (VAE) pretrained with force time series reconstruction. 
By projecting force latent representations and short state history as additional conditioning tokens to the action expert module, we enable VLAs to leverage accumulated contact event history to guide manipulation. 
We evaluate \name{} on three memory-dependent tasks, including finding a hidden block, pressing a button, and wiping a dish for a specific number of times. 
Our lightweight force memory achieves over 80\% success rate with minimal inference overhead, significantly outperforming baseline approaches.
Project page: \url{https://qft-333.github.io/FM-VLA-Page/}.
\end{abstract}


\section{Introduction}
\label{sec:intro}

Vision-language-action (VLA) models~\citep{brohan2023rt2, kim2024openvla, black2025pi0, intelligence2025pi05} have achieved remarkable generalization across diverse manipulation tasks by leveraging internet-scale vision-language pretraining. 
However, most existing VLA architectures rely solely on the current observation, treating decision-making as a memoryless mapping from the current state to future action. 
While this Markovian assumption suffices for many tasks, it becomes inadequate in real-world scenarios where decisions depend on long-horizon temporal context rather than the instantaneous observation alone.
For example, tasks that require counting repeated actions, tracking interaction progress over multiple steps, or reasoning over partially observed states (e.g., searching for a hidden object) are inherently non-Markovian, as the correct action often depends on past environmental and agent states.
 
Prior work has explored vision-based memory mechanisms to address this challenge. 
MemoryVLA~\citep{shi2025memoryvla} maintains an explicit memory bank over past observations, while MEM~\citep{torne2026mem} models multi-scale memory spanning long-term language context and short-term visual histories. However, these approaches still face several key limitations.
First, they struggle with tasks where visual changes are subtle or observations are severely occluded. For instance, repeated button presses with small motions may result in no observable change in the visual input.
Second, visual memory can introduce additional computational overhead, as storing and attending to past frames increases input token length and inference latency. As a result, vision-based memory alone can be less reliable and efficient in such scenarios, particularly under weak or occluded visual feedback.

In contrast, many interaction-relevant signals such as contact events, force magnitude, and the count of repeated actions, are naturally captured by force sensor measurements. These signals provide a more direct and unambiguous representation of interaction dynamics.
Recent works have also incorporated force or torque sensing into VLA models, such as ForceVLA~\citep{yu2025forcevla} and TA-VLA~\citep{zhang2025tavla}. However, their primary focus is on using force signals to improve action prediction based on current observations or short-term context, rather than leveraging force as a mechanism for tracking interaction progress. As a result, they can capture local interaction states such as whether contact is established or how much force is being applied, but fail to accumulate the long-horizon temporal information required for non-Markovian decision-making.

To bridge this gap, we present \name{}, a vision-language-action model that incorporates \textbf{force-based temporal memory} for long-horizon, contact-rich tasks. 
The key challenge lies in transforming long-term force history into meaningful representations that effectively guide action generation, while the raw force signals are high-frequency, noisy, and unstructured.
An intuitive baseline is to employ a temporally compressive network to inject full-trajectory force signals into the VLA action expert and learn end-to-end conditioning, as in~\cite{zhang2025tavla}. However, we find that such an approach struggles to learn effective representations for guiding action generation over extended horizons. 

To address this, we adopt a two-stage paradigm. 
We first pre-train a Variational Autoencoder (VAE)~\cite{diederik2019introduction, jaegle2022perceiveriogeneralarchitecture} to compress the noisy force history into a compact yet informative latent space, which is subsequently used to condition the action expert. Such a design distills a lightweight yet expressive summary of accumulated contact events from long-horizon, noisy force sequences without requiring textual or class labels. 
Consequently, the action expert can more effectively leverage the historical force context to produce precise and coherent future actions for contact-rich, memory-dependent tasks.
We further observe that conditioning solely on force history may induce undesirable repetitive behaviors before contact due to the lack of short-term motion awareness. 
To mitigate this artifact, we introduce a lightweight projector that incorporates a short window of proprioceptive state as an additional conditioning signal.
We show that \name{} effectively leverages force history to solve memory-dependent manipulation tasks that would otherwise fail, achieving significantly higher success rates than visual-memory-based approaches while maintaining inference-time efficiency. This suggests a promising direction for more capable VLA systems in complex real-world manipulations.

\textbf{Our contributions are summarized as follows:}
\vspace{-9pt}
\begin{itemize}
    \item We present \name{}, the first VLA model with force-based memory, enabling temporal reasoning over force and joint state histories for non-Markovian, contact-rich manipulation.
    
    \item We introduce a lightweight VAE to compress long-horizon force signals into a compact, informative representation for accurate action generation, significantly outperforming naive end-to-end injection baselines.
    
    \item We evaluate with a bimanual robot on three contact-rich tasks and show that force memory enables temporal context understanding, e.g., counting contact events, outperforming both memoryless and vision-based memory baselines by a large margin.
\end{itemize}

\begin{figure}[t]
    \centering
    \includegraphics[width=0.98\textwidth]{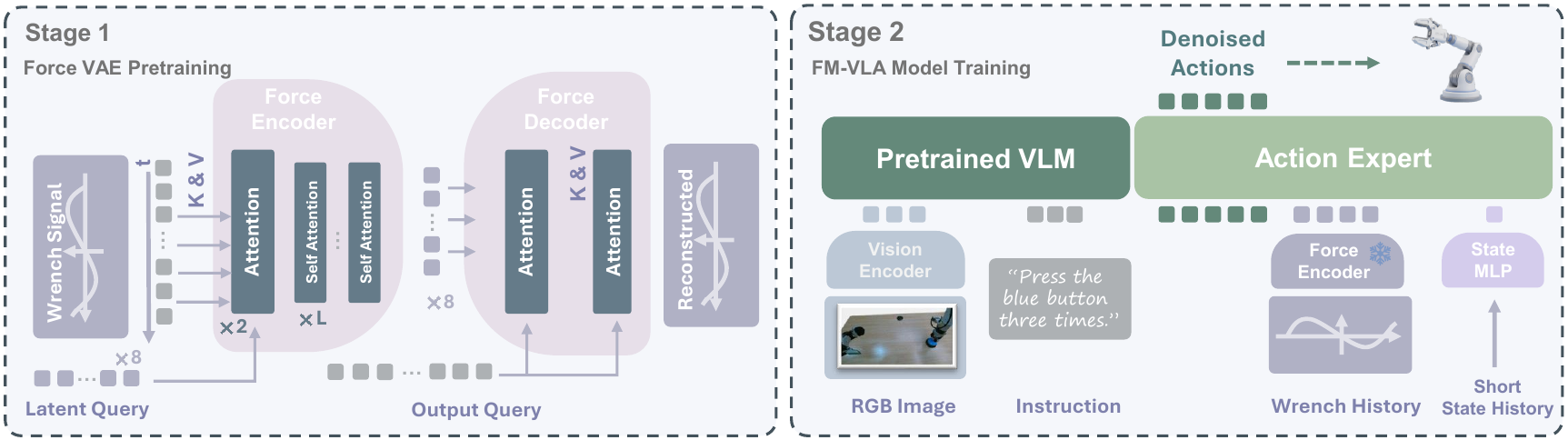}
    \caption{\textbf{Overview.} We augment a VLA with lightweight force-based temporal memory. Force/torque histories are encoded by a VAE encoder into latent representations, projected via MLP into force memory tokens that condition the flow matching action expert. A short-term state memory token is further appended to provide motion context.}
    \label{fig:overview}
    \vspace{-4pt}
\end{figure}

\section{Related Work}
\label{sec:related}
\paravspace
\paragraph{Vision-language-action models.}

Vision-language-action (VLA) models~\cite{brohan2023rt2} build on large pre-trained
vision-language backbones~\cite{steiner2024paligemma,comanici2025gemini,bai2025qwen3} to produce end-to-end visuomotor policies. Recent works in this line have explored a range of improvements, including network architectures, action modeling and generation, more complex data compositions, and learning objectives~\cite{kim2024openvla,black2025pi0,intelligence2025pi05,liu2024rdt,geminirobotics2025,li2024cogact,bjorck2025gr00t,qu2025spatialvla,cheang2025gr}.
However, these policies remain \emph{Markovian}, mapping the current observation directly to the next action chunk without maintaining explicit memory of past states.

\paravspace
\paragraph{Memory in VLA models.}
A growing body of work moves beyond the memoryless assumption, but overwhelmingly along two axes: language and vision. 
 
MemoryVLA~\cite{shi2025memoryvla} maintains a perceptual-cognitive memory bank from which a diffusion action expert retrieves and fuses past visual tokens.
MEM~\cite{torne2026mem} further combines a short-term video memory, which encodes recent tens of frames (around 1 second per frame), and a long-term text-based summary for long-horizon tasks. 
A broader family of follow-ups explores alternative visual-memory designs, all of which operate on \emph{visual} or \emph{linguistic} tokens~\cite{li2026rememvla,lin2025echovla,lei2026vpwem,wang2026vqmemory,sun2026tempofit,haresh2026notes,li2026dualmem,chung2026objectmem}. 
These approaches primarily encode information derived from visual observations or language, leaving interaction-grounded signals from prior contact largely unexplored. 
This limitation becomes critical when task-relevant state changes are not visually observable but are reflected in physical interaction.

\paravspace
\paragraph{Force-augmented VLAs.}
A recent line of work explores the integration of force or tactile signals into VLA models.
TA-VLA~\cite{zhang2025tavla} systematically charts the design space of torque-aware VLAs and proposes an action decoder-side torque adapter.

ForceVLA~\cite{yu2025forcevla} similarly fuses the 6-axis wrench as a first-class modality with vision-language tokens via a Mixture-of-Experts design.

A growing body of follow-up work further advances this direction by exploring force modality integration, including flow-matching control, contact-gated routing, adaptive fusion, and world-model-style conditioning~\cite{zhang2026forceflow,zhang2026tacvla,gubernatorov2026hapticvla,li2026atvla,huang2026tafvla,zhang2025vtla,ye2025dreamtacvla}.
While these methods show that contact information enables challenging fine-grained manipulation, they condition only on a \emph{short window} of force readings concurrent with the current action. This provides a rich corrective signal for the immediate motion, but no memory of prior interactions over an episode, such as how many times a button has been pressed.

Our work is, to our knowledge, the first to treat the proprioceptive--wrench stream as a \emph{lightweight, long-horizon memory} for VLAs, rather than as instantaneous conditioning.

\section{Methodology}
\label{sec:method}
 
\subsection{Problem Formulation}

We consider learning a manipulation policy $\pi(a_t \mid o_t, l, h_t)$ that maps the current observation $o_t$, a language instruction $l$, and a temporal history $h_t$ to an action $a_t$. Vanilla VLA models \cite{brohan2023rt2, kim2024openvla,black2025pi0,intelligence2025pi05,bjorck2025gr00t} operate as memoryless policies $\pi(a_t \mid o_t, l)$, which suffices for single-step tasks but fails when the correct action depends on past interactions or other history.

Our method augments the policy with two complementary proprioceptive streams. 
The first is a \emph{long-horizon} wrench history $\{f_\tau\}_{\tau=1}^{t}$, where each $f_\tau \in \mathbb{R}^{d_f}$ with $d_f=6$ stacks the 3-axis force and 3-axis torque measured by a wrist-mounted six-axis force/torque (F/T) sensor; this stream captures accumulated contact events over an entire episode, providing temporal context for policy decision-making. 
The second is a \emph{short-window} joint-state history $\{s_\tau\}_{\tau=t-W+1}^{t}$, where each $s_\tau \in \mathbb{R}^{d_s}$ concatenates the joint positions of all arms with the gripper state.
$d_s$ depends on the robot embodiment, e.g., $d_s = 16$ for a 7-DoF bimanual setup with two 1-D grippers; this stream captures the most recent proprioceptive dynamics, complementing force history and mitigating repetitive action behaviors before contact. 
Both signals are low-dimensional, available at high frequency, and directly encode contact dynamics that may be invisible or ambiguous in visual observations.

Given the two streams introduced above, we encode them into a temporal-history representation $h_t$ used in $\pi(a_t \mid o_t, l, h_t)$ as follows.
\begin{equation}
\label{eq:h_t}
    h_t \;=\; \big[\, \underbrace{\mathrm{Enc}_\phi\!\big(\{f_\tau\}_{\tau=1}^{t}\big)}_{\text{wrench history } Z_f \in \mathbb{R}^{K \times d_h}} \;\;\Vert\;\; \underbrace{\mathrm{Proj}_\psi\!\big(\{s_\tau\}_{\tau=t-W+1}^{t}\big)}_{\text{state history window} z_s \in \mathbb{R}^{d_h}} \,\big],
\end{equation}
where $\mathrm{Enc}_\phi$ is a pretrained VAE encoder that compresses the unbounded wrench history into a fixed set of $K$ latent tokens with dimension $d_h$,

$\mathrm{Proj}_\psi$ is a lightweight linear projection that maps the joint-state window $\{s_\tau\}_{\tau=t-W+1}^{t}$ to a single token, learned end-to-end with the VLA finetuning, and $\Vert$ is context-wise concatenation.
We append both groups of tokens to the noisy-action tokens of the flow-matching action expert, realizing the memory-augmented policy of our problem formulation.

\subsection{\name{} Architecture}
Figure~\ref{fig:overview} provides an overview of \name{} architecture.
We build \name{} based on $\pi_{0.5}$~\citep{intelligence2025pi05}, which consists of a VLM (PaliGemma~\citep{beyer2024paligemma} with a SigLIP~\citep{zhai2023sigmoid} vision encoder), and a flow-matching~\cite{lipman2022flow} action expert.
The VLM processes the current image and language instruction, and its internal features condition the flow-matching action expert via cross-attention to generate action chunks.

To incorporate force-based temporal memory, we first process raw wrench readings, then encode long-horizon wrench history and short-window state history through two specialized modules.
The wrench stream goes through a frozen VAE encoder pretrained with reconstruction; the state stream goes through a single linear projection learned end-to-end with the policy. 
The wrench encoder yields $K$ tokens $Z_f$ and the state encoder yields a single token $z_s$, which are appended to the action-expert suffix, after the noisy-action tokens.
We show our full training pipeline in Section~\ref{sec:training}.

\subsubsection{Wrench History Processing}
\label{sec:history_buffer}

\paragraph{First-order EMA smoothing.} 
Raw wrench readings are noisy and contain high-frequency content that does not contain task-relevant information.
We apply a causal first-order exponential moving average $\tilde{f}_\tau = \alpha f_\tau + (1-\alpha)\tilde{f}_{\tau-1}$, which removes most noise while preserving onsets and peaks.

\paravspace
\paragraph{Randomized noise pre-padding.}

We notice the length of the wrench history leaks temporal episode progress, letting the model shortcut on sequence length for action generation rather than utilizing signal temporal structure.
To remove this cue, during training we prepend each history with a random-length prefix of low-amplitude Gaussian noise (uniformly sampled up to 10\,s).
This randomizes the padding--signal boundary and forces the encoder to rely on signal content. The augmentation is disabled at inference.

\subsubsection{Proprioceptive Memory Design}
\label{sec:temporal_memory}

\paragraph{Force memory encoder.}
We train a VAE on wrench time-series reconstruction as a task-agnostic encoder, producing a structured latent space that captures temporal patterns in contact events such as force magnitudes, contact counts, without requiring task-specific labels.
Both encoder and decoder are instantiated as Perceiver-IO~\citep{jaegle2022perceiveriogeneralarchitecture} stacks operating on a small set of $K$ learnable latent query tokens. 
Given a wrench history $F = [f_1, \ldots, f_T] \in \mathbb{R}^{T \times d_f}$, the signal $f_t$ at each time-step is first quantile-normalized based on the statistics of the entire dataset.
These normalized signals are subsequently projected through an input MLP and integrated with a Fourier positional encoding to yield wrench tokens.

The encoder leverages cross-attention to extract wrench features into the latent queries, interleaved with self-attention blocks over the latents. Finally, a per-latent linear head projects each encoded latents into the posterior parameters:
\begin{equation}
    (\mu_k, \log\sigma_k^2) = \text{Head}_\text{VAE}\big(\text{Enc}_\phi(F)_k\big), \quad z_k = \mu_k + \sigma_k \odot \epsilon_k, \;\; \epsilon_k \sim \mathcal{N}(0, I),
\end{equation}
yielding a latent representation $Z \in \mathbb{R}^{K \times d_z}$ of $K$ tokens. The decoder reverses this process via cross-attention layers, where time-step-specific Fourier-encoded queries attend to the latent tokens $Z$ to produce the reconstructed sequence $\hat{F} \in \mathbb{R}^{T \times d_f}$.

\paravspace
\paragraph{Short state history.}
Different from the wrench stream, the proprioceptive role of the joint-state stream is well captured by the very recent states.
The action expert only needs to know ``where the arms are and where they are heading''. 
We therefore avoid a second VAE on the state side and use a lightweight projection layer with no pretraining.
Concretely, at each control step $t$ we extract a short window $S_t \in \mathbb{R}^{W \times d_s}$ of the most recent joint-state frames by sub-sampling the state stream at a fixed stride. This window covers the last second of motion. We flatten $S_t$ and project it to a single state history token with a single zero-initialized linear layer.

\subsubsection{Memory Token Injection}
\label{sec:injection}

Both proprioceptive memory streams enter the policy exclusively through the action-expert \emph{suffix}, after the noisy-action tokens. From the frozen force encoder we take only the posterior mean $\mu_f \in \mathbb{R}^{K \times d_z}$ and project from the VAE latent dimension $d_z$ to the action-expert hidden dimension $d_h$ using a zero-initialized linear layer, yielding wrench memory tokens $Z_f \in \mathbb{R}^{K \times d_h}$.
Together with the single state token $z_s$, the action-expert sequence has the layout
\begin{equation}
    \underbrace{[\, a_k^{(1)}, \ldots, a_k^{(H)} \,]}_{\text{noisy-action tokens}} \;\Vert\; \underbrace{[\, Z_f^{(1)}, \ldots, Z_f^{(K)} \,]}_{\text{wrench memory}} \;\Vert\; \underbrace{[\, z_s \,]}_{\text{state window}}.
\end{equation}
The wrench tokens are appended immediately after the noisy actions and the state-window token is appended last. 
Placing both memory streams in the post-position keeps the noisy-action tokens at the same RoPE positions they had during base-policy pretraining.

\subsection{Training}
\label{sec:training}

The training of \name{} proceeds in two stages: we first learn a task-agnostic representation of temporal dynamics with a VAE for force signals, and then integrate this representation into the VLA policy through fine-tuning.

\paravspace
\paragraph{Stage 1: Force-VAE pretraining.} 
We train the force memory VAE (Section~\ref{sec:temporal_memory}) on the wrench histories of all tasks jointly with the masked-ELBO objective $\mathcal{L}_{\text{VAE}}$. This stage is task-agnostic and produces a general-purpose wrench latent space. 

The VAE is trained with a masked reconstruction term over valid frames and a free-bits-regularized KL on each latent dimension,
\begin{equation}
    \mathcal{L}_{\text{VAE}} = \frac{1}{\sum_\tau m_\tau \cdot d_f}\sum_{\tau=1}^{T} m_\tau \|f_\tau - \hat{f}_\tau\|^2 \;+\; \beta \cdot \frac{1}{K d_z}\sum_{k,j} \max\!\big(D_\text{KL}^{(k,j)}, \lambda\big),
\end{equation}
where $m_\tau \in \{0,1\}$ masks padding frames, $D_\text{KL}^{(k,j)}$ is the per-dimension KL of $\mathcal{N}(\mu_{k,j}, \sigma_{k,j}^2)$ against the standard normal prior, $\beta$ controls the regularization strength, and the per-dimension free-bits floor $\lambda$ prevents posterior collapse by switching off the KL gradient on dimensions that already encode less than $\lambda$ nats. We pretrain a single VAE jointly on the wrench histories of all task subsets with inverse-frequency task sampling, so that small datasets are not drowned out by larger ones.

\paravspace
\paragraph{Stage 2: VLA finetuning.} 
We freeze the force encoder and switch it to evaluation mode for fine-tuning on our dataset. 
Only the posterior mean $\mu_f$ is used in the forward pass, without reparameterization noise. Following the rectified-flow recipe~\cite{liu2022flow} of $\pi_{0.5}$, we sample a clean action chunk $a_0$ from the dataset, Gaussian noise $\epsilon \sim \mathcal{N}(0, I)$, and a noise level $k \in [0,1]$, and form the interpolated sample $a_k = (1-k)\,a_0 + k\,\epsilon$. We jointly fine-tune the VLM encoder, the flow-matching action expert, the short-window state projector $\mathrm{Proj}_\psi$, and the wrench latent projector $W_f$ to predict the constant velocity $\epsilon - a_0$ of the straight-line path, with $v_\theta$ now additionally conditioned on the wrench memory and state tokens:
\begin{equation}
    \mathcal{L} = \mathbb{E}_{a_0,\, \epsilon,\, k}\left[\big\|v_\theta(a_k, k, c_t, Z_f, z_s) - (\epsilon - a_0)\big\|^2\right],
\end{equation}
where $Z_f$ and $z_s$ are appended to the noisy action chunk in the order described in Section~\ref{sec:injection}.

\section{Experiments}
\label{sec:experiments}

\subsection{Experimental Setup}

\paragraph{Robot platform.}
We collect datasets and conduct experiments on an AgiBot G1 bimanual humanoid robot~\citep{agibot2024}, which features two 7-DoF arms with two 1-Dof grippers as end-effectors. Each wrist of the end-effectors is equipped with a 6-axis wrench sensor including 3-axis force and 3-axis torque at 100\,Hz, making the platform well-suited for contact-rich manipulation. The policy input consists of the 6-DoF wrist wrench, three RGB streams from the head and two wrist cameras, and a proprioceptive state vector including arm joint positions and gripper states.

\begin{figure}[t]
    \centering
    \includegraphics[width=0.95\textwidth]{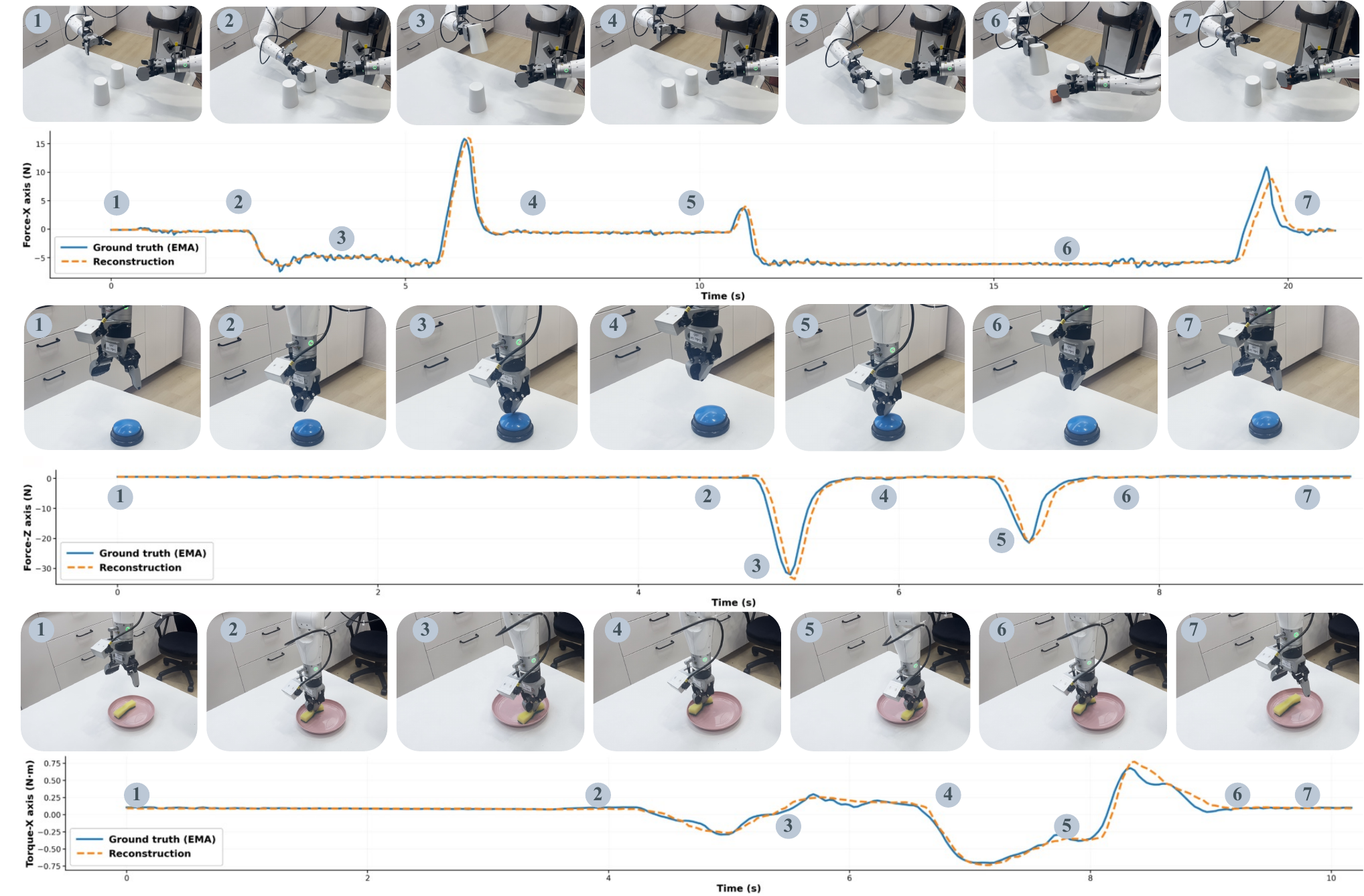}
    \vspace{-2pt}
    \caption{\textbf{Trajectory and force signal visualization.} For each task, we visualize the \name{} inference trajectory and a selected channel of force readings, with each frame's time marked. \name{} correctly memorizes contact events and complete manipulation successfully.}
    \label{fig:trajectory_force}
    \vspace{-6pt}
\end{figure}

\paravspace
\paragraph{Task suite.}
We design three contact-rich bimanual tasks (\Cref{fig:trajectory_force}) that require integrating temporal history: 
(1) \emph{Find a Block Under Two Cups}: The robot must sequentially lift two upside-down, visually identical cups to pick up a hidden wooden block. Since the scene returns to its original appearance after a cup is placed back, the policy must remember which cup has already been inspected.
(2) \emph{Push Buttons}: The robot is instructed to press a button exactly $N \in \{1,2,3\}$ times and stop. Due to the button's minimal travel distance, each press yields negligible visual displacement but generates a distinct, sharp wrench impulse. Thus, counting completed presses fundamentally depends on accumulating contact events over time.
(3) \emph{Wipe Dishes}: The robot grasps a sponge and wipes the interior of a bowl for a specified number of passes $N \in \{1,2,3\}$. Similar to the button task, the visual scene changes only marginally, making wrench history the dominant cue for tracking interaction progress.
We collect 200, 350, and 200 teleoperated demonstrations for Task 1, Task 2, and Task 3, respectively, using a VR-based control interface to construct the training dataset.

\paravspace
\paragraph{Baselines.}
We compare \name{} against three baselines to evaluate different memory paradigms:
(1) \textbf{$\mathbf{\pi_{0.5}}$} is the base VLA without temporal conditioning, establishing a memoryless (Markovian) reference.
(2) \textbf{TA-VLA}~\citep{zhang2025tavla} conditions the action expert on a short moving window of recent wrench readings, representing prior force-based approaches.
(3) \textbf{$\pi$-MEM} is our reimplementation of the state-of-the-art visual memory mechanism from MEM~\citep{torne2026mem} on top of $\pi_{0.5}$ base model, testing whether purely visual temporal context suffices for visually ambiguous, contact-driven tasks.

\subsection{Main Results}

\begin{table}[t]
\centering
\caption{\textbf{Main results.} Success rates (\%) of different methods on all tasks.}
\label{tab:main}
\begin{tabular}{lccc|c}
\toprule
\textbf{Method} & \textbf{Task 1: Cups} & \textbf{Task 2: Buttons} & \textbf{Task 3: Wipe} & \textbf{Average} \\
\midrule
$\pi_{0.5}$ (no history) & ~~72.2 (13/18) & 11.1 (2/18) & ~~0.0 (0/18) & 27.8 \\
TA-VLA~\citep{zhang2025tavla} & ~~50.0 (~~9/18) & 11.1 (2/18) & ~~5.6 (1/18) & 22.2 \\
$\pi$-MEM~\citep{torne2026mem} & ~~77.8 (14/18) & 33.3 (6/18) & 50.0 (9/18) & 53.7 \\
\midrule
\multicolumn{5}{l}{\emph{Modality ablation (VAE)}} \\
Force history only & ~~55.6 (10/18) & ~~0.0 (0/18) & 22.2 (4/18) & 25.9 \\
State history only & \textbf{100.0 (18/18)} & 11.1 (2/18) & 11.1 (2/18) & 40.7 \\
\midrule
\multicolumn{5}{l}{\emph{Architecture ablation (Force + Short State History)}} \\
FM-VLA (GRU) & ~~55.6 (10/18) & 38.9 (~~7/18) & ~~5.6 (~~1/18) & 33.3 \\
FM-VLA (Q-Former) & \textbf{100.0 (18/18)} & 16.7 (~~3/18) & 55.6 (10/18) & 57.4 \\
\textbf{FM-VLA (VAE, ours)} & \textbf{100.0 (18/18)} & \textbf{72.2 (13/18)} & \textbf{77.8 (14/18)} & \textbf{83.3} \\
\bottomrule
\end{tabular}
\vspace{-4pt}
\end{table}

\paragraph{Force memory effectiveness.}
We use the success rate as the primary metric and consider a trial successful if the robot completes the task and reaches a stable termination (opened gripper, no motion for $\sim$3\,s).
\name{} consistently outperforms all baselines, achieving an average success rate of 83.3\%. 
The advantage is particularly pronounced on the \emph{Buttons} and \emph{Wipe} tasks, which require precise temporal reasoning over force signals. 
By contrast, the memoryless $\pi_{0.5}$ struggles significantly (27.8\% average), failing entirely on the wiping task. 
TA-VLA performs similarly poorly; its short moving window of force captures instantaneous contact but cannot retain the long-term event counts required for these tasks.
Notably, while the visual-memory baseline $\pi$-MEM shows moderate improvements on \emph{Cups} and \emph{Wipe}, it fails heavily on the \emph{Buttons} task (33.3\% vs. our 72.2\%). This confirms that visual memory is insufficient for tasks lacking clear visual state changes.

\subsection{Ablation Studies}
To understand the contribution of each component in \name{}, we conduct extensive ablations on input modalities and architectural designs.

\paravspace
\paragraph{Modality: What to remember?}
We isolate the contribution of force and state histories. \textbf{Force-only} utilizes only the force memory tokens from the VAE, while \textbf{State-only} relies purely on proprioceptive joint states. 
As shown in \Cref{tab:main}, utilizing either modality alone leads to severe performance degradation. Force-only drops to 25.9\% on average because the policy lacks short-term spatial awareness before making contact, leading to erratic pre-contact motions. Conversely, State-only performs perfectly on the spatially-driven \emph{Cups} task but fails on contact-driven counting tasks. This demonstrates that combining long-term force memory with short-term state history is critical for stable, contact-rich manipulation.

\paravspace
\paragraph{Architecture: Why VAE?}
Compressing noisy, high-frequency force histories into compact tokens is the key design challenge. We compare our VAE encoder against a \textbf{GRU}~\cite{DBLP:journals/corr/ChungGCB14} recurrent encoder and a \textbf{Q-Former}~\citep{li2023blip2} cross-attention module, and both alternatives significantly underperform. The GRU suffers from vanishing gradients over long 100\,Hz sequences and loses early contact events (e.g., 5.6\% on \emph{Wipe}), while the Q-Former overfits to instantaneous peaks instead of the holistic temporal structure.
By contrast, our VAE is pretrained on a continuous wrench reconstruction objective, which forces the latent space to encode macroscopic structure, e.g., force magnitudes, onset timings, and contact counts, in a few tokens, making task-relevant signals easy for the VLA action expert to extract.

\begin{figure}[t]
\centering
\begin{minipage}[b]{0.53\textwidth}
\centering
\captionof{table}{\textbf{Inference efficiency.} \name{} adds minimal overhead vs. $\pi$-MEM (K frame vision memory).}
\label{tab:speed}
\vspace{2pt}
\setlength{\tabcolsep}{4pt}
\small
\begin{tabular}{lcc}
\toprule
\textbf{Method} & \textbf{Latency (ms)} & \textbf{$\Delta$ vs.\ base (ms)} \\
\midrule
$\pi_{0.5}$ (base) & ~~60.7 $\pm$ 0.3 & --- \\
$\pi$-MEM~\citep{torne2026mem} ($K{=}5$) & ~~99.8 $\pm$ 0.4 & +39.1 \\
$\pi$-MEM~\citep{torne2026mem} ($K{=}16$) & 190.0 $\pm$ 1.0 & +129.3 \\
\textbf{\name{} (ours)} & ~~64.0 $\pm$ 0.4 & \textbf{+3.3} \\
\bottomrule
\end{tabular}
\vspace{0.6em}
\end{minipage}\hfill
\begin{minipage}[b]{0.42\textwidth}
\centering
\setlength{\abovecaptionskip}{0pt}
\includegraphics[width=\linewidth]{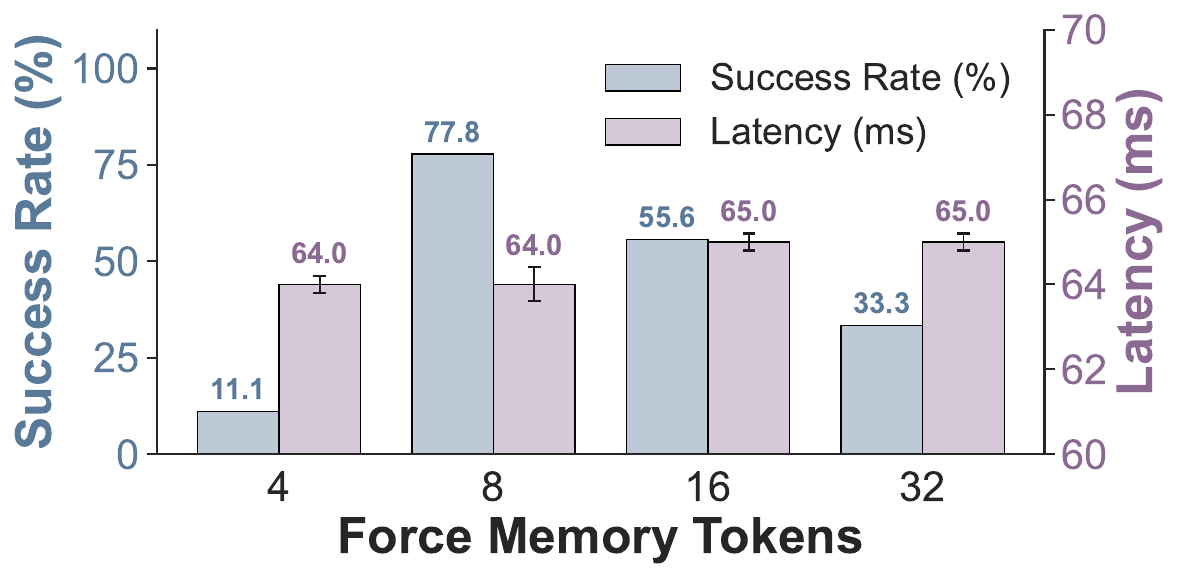}
\captionof{figure}{\textbf{Token count ablation.} } 
\label{fig:tokens}
\end{minipage}
\vspace{-4pt}
\end{figure}

\paravspace
\paragraph{Capacity: How many tokens?} 
We ablate the VAE latent token count in $\{4, 8, 16, 32\}$ on \emph{Wipe Dishes} (\Cref{fig:tokens}). 
4 tokens form an informational bottleneck, while 16 and 32 tokens unexpectedly degrade performance. 
We attribute this to distribution shift, as the pretrained $\pi_{0.5}$ action expert observes at most 50 tokens during training, so 32 extra force tokens exceed this limit and disrupt coherent action generation.
We use 8 tokens as it delivers the peak success rate.

\subsection{Inference Efficiency}
We compare \name{} with the base policy $\pi_{0.5}$ and vision-based memory in terms of inference latency on an RTX 4090, as shown in \Cref{tab:speed}.
\name{} shows a 64~ms latency, introducing negligible 3~ms overhead on top of the base model,
In comparison, $\pi$-MEM shows a 100~ms inference latency and 39~ms increase, due to the overhead of multiple RGB frames input to the vision/video encoder as memory. This latency can scale further to near 190~ms, which agrees with \citep{torne2026mem}.
This comparison highlights the efficiency advantage of \name{} over vision-based memory.

\section{Limitations}
Our current VAE latent space introduces a fixed bottleneck of 8 tokens; for very long-horizon tasks requiring memory over hundreds of contact events, hierarchical or adaptive compression may be needed. We train the VAE on force data from the demonstration dataset; pretraining on large-scale robot datasets with diverse force/torque recordings could further improve the performance.

\section{Conclusion}
\label{sec:conclusion}
We presented \name{}, a VLA with force-based memory that enables temporal context reasoning over contact history. 
To achieve task-agnostic wrench signal understanding, we trained a VAE on force/torque time-series reconstruction and injected the frozen encoder's latent representations as force memory tokens into the action expert.
This equips the model with accumulated event memory even when the visual observation is limited or ambiguous, a capability neither vision-based memory nor single-token force conditioning provides. 
We evaluated on three contact-rich tasks on a bimanual robot that require temporal event reasoning and showed that \name{} outperforms baseline methods by a significant margin while introducing negligible additional inference overhead.

\section{Acknowledgments}
This work was supported by the Beijing Natural Science Foundation under Grant No. QY25049.

\bibliography{references}
\clearpage
\appendix

\section{Additional Real-World Results}
Additional real-world inference results of our method can be found in the supplementary video.
\section{Task Definitions and Evaluation Protocol}
\label{app:tasks}

This appendix details the per-task rules used during data collection and
evaluation, the success criteria used to compute the success rates reported in
the main text, and the natural-language instructions provided to the policy.

\subsection{Task Rules}
\label{app:task_rules}

\paragraph{Task~1 — Find a Block Under Two Cups.}
Two visually identical, opaque cups are placed upside-down on the table in
front of the robot. A small wooden block is hidden under exactly one of them,
out of the robot's view. The robot must inspect the cups from front to back:
it lifts the cup closer to itself with the \emph{right} arm; if the block is
revealed it picks up the block with the \emph{left} arm; if no block is
revealed it places the cup back on the table and proceeds to the second cup.
Once the block is grasped by the left hand, the robot returns the cup
currently held in the right hand to the table. The robot is not allowed to
re-inspect a cup that has already been lifted, so completing the task on the
second cup requires remembering which cup has already been checked.

\paragraph{Task~2 — Press a Blue Button.}
A blue button is placed on the table. The instruction specifies
a target count $N \in \{1, 2, 3\}$. The robot closes its right-hand gripper
and presses the button $N$ times, then opens the gripper to signal task
completion. A press only counts when the gripper compresses the button deep
enough to trigger its built-in click, which provides an audible feedback
signal; soft taps that do not click are not counted toward $N$. Because each
mechanical press produces almost no visible displacement, the count must be
inferred from the wrist wrench history.

\paragraph{Task~3 — Wipe a Bowl with a Sponge.}
A sponge and a shallow bowl are placed on the table. The instruction specifies
a target number of wiping rounds $N \in \{1, 2, 3\}$, where one round is a
single back-and-forth pass across the interior of the bowl. The robot grasps
the sponge with the right hand and wipes the bowl for $N$ rounds, then opens
its gripper to release the sponge and stop. During each round the sponge must
maintain contact with the bowl from the left rim to the right rim without
losing contact in the middle; partial passes do not count toward $N$.

\subsection{Success Criteria and Evaluation Counts}
\label{app:success}

For every method we report success rate over 18 trials per task, broken down
as follows.

\paragraph{Task~1.}
We run 9 trials with the block hidden under the front cup and 9 trials with
the block under the back cup. A trial is considered successful when
\emph{all} of the following hold:
(i) the robot lifts each cup at most once (no re-inspection of a previously
opened cup);
(ii) the block ends up grasped in the left hand;
(iii) the cup that was lifted last is placed back on the table by the right
hand.

\paragraph{Task~2.}
We run 6 trials for each of $N \in \{1, 2, 3\}$, for a total of 18 trials. A
trial is considered successful when the robot produces exactly $N$ audible
clicks and then opens the gripper. Taps that do not trigger an audible click
are not counted toward $N$; both undercounting and overcounting are recorded
as failures.

\paragraph{Task~3.}
We run 6 trials for each of $N \in \{1, 2, 3\}$, for a total of 18 trials. A
trial is considered successful when the robot completes exactly $N$ full
back-and-forth wiping rounds, where each round spans from the left rim of the
bowl to the right rim without breaking contact in the middle, after which the
robot opens its gripper to release the sponge.

\subsection{Language Instructions}
\label{app:instructions}

The natural-language instruction provided to the policy at inference time is
fixed per task and per target count $N$. The exact strings used in our
evaluation are:

\paragraph{Task~1.}
\begin{itemize}
    \item \texttt{Find the block under the cup and pick it up with left hand.}
\end{itemize}

\paragraph{Task~2.}
\begin{itemize}
    \item \texttt{Press the blue button once.}
    \item \texttt{Press the blue button twice.}
    \item \texttt{Press the blue button three times.}
\end{itemize}

\paragraph{Task~3.}
\begin{itemize}
    \item \texttt{Wipe the bowl with a sponge back and forth for one round.}
    \item \texttt{Wipe the bowl with a sponge back and forth for two rounds.}
    \item \texttt{Wipe the bowl with a sponge back and forth for three rounds.}
\end{itemize}
\section{Model Initialization}
We initialize the model from the publicly released $\pi_{0.5}$ checkpoint provided by the OpenPI framework and further pretrain it on the Zhiyuan Challenge dataset. The model is trained for 150K steps with a batch size of 64, and delta actions are used as the prediction target. All experiments in this paper are trained starting from this checkpoint.
\section{Baseline Implementation Details}
\label{app:baselines}

All baselines are post-trained on the same dataset as
\name{} (Task~1: 200 demos, Task~2: 350 demos, Task~3: 200 demos) using the same normalized statistics, the same learning-rate schedule (1k warm-up, peak $5\!\times\!10^{-5}$, no decay phase, 50k total steps), the same global batch
size of 32, and the same per-view image dropout with $p\!=\!0.4$.

\paragraph{$\pi_{0.5}$ (no history).}
We use the original $\pi_{0.5}$ architecture without modification. The policy conditions on
the current head camera, two wrist cameras, the language instruction, and the
current proprioceptive state, and predicts a 30-step action chunk via flow
matching. No wrench, no force-history, and no visual-history conditioning is
added; this is a pure single-frame VLA baseline post-trained on our
demonstrations.

\paragraph{TA-VLA.}
The original TA-VLA paper builds its torque-aware
adapter on top of $\pi_{0}$. For a fair comparison with \name{}, our TA-VLA
baseline is a $\pi_{0.5}$-based reimplementation that adopts the two design
choices reported as best in the paper: (i) a short window of
recent wrench readings is injected as an additional token into the action
expert \emph{before} the noisy-action tokens, and (ii) an
auxiliary \emph{future force prediction} head shares the flow-matching
backbone with the action head. Concretely, at each control step we sample a
short window of 10 right-wrist 6-axis wrench frames at offsets
$\{-27, -24, \ldots, -3, 0\}$ relative to the current step (stride 3 at
30\,Hz, spanning roughly the last 0.9\,s), flatten the resulting
$10 \times 6$ tensor, project it through a single linear layer to the
action-expert hidden width, and prepend the resulting token to the noisy
action chunk. Future force prediction is implemented by extending the flow
matching target from a $32$-dim action vector to a $32+6 = 38$-dim vector that concatenates the action chunk with the corresponding future wrench chunk; both heads are trained jointly with an action loss weight of
$1.0$ and a force-prediction loss weight of $0.1$. The VLM, action expert,
short-force projector, and force-prediction head are all fine-tuned
end-to-end.

\paragraph{$\pi$-MEM.}
We
adopt MEM's space-time separable attention design: after the spatial
self-attention of every 4th SigLIP encoder layer, we insert a
\emph{causal temporal} self-attention sub-block that operates along the
per-patch time axis across the $K$ history frames. To preserve the SigLIP
pretraining, this temporal sub-block adds \emph{no} new vision parameters:
it reuses the same layer's pre-attention LayerNorm and the same
$W_Q\!/\!W_K\!/\!W_V\!/\!W_O$ projections as the spatial attention, and
applies a fixed sinusoidal temporal positional embedding constructed so that
the embedding at the current timestep is exactly zero. After the encoder,
only the patch tokens of the current frame are forwarded to PaliGemma, so
the language backbone and action expert see exactly the same number and
shape of vision tokens as the single-frame $\pi_{0.5}$ baseline.

At training and inference, the data loader samples $K\!=\!5$ history frames
per camera at a stride of $45$ dataset frames ($\approx 1.5\,$s at $30\,$Hz)
ending at the current timestep, for each of the three cameras (head,
left-wrist, right-wrist); the most recent frame is exactly the current
observation, so $\pi$-MEM is byte-identical to $\pi_{0.5}$ when $K\!=\!1$.
To isolate the marginal contribution of visual history, $\pi$-MEM does
\emph{not} use any wrench injection and does \emph{not} consume the
proprioceptive state-history window used by \name{}. 

\section{Implementation Details of \name{}}
\label{app:impl}

This appendix complements the experiments part with the exact
hyperparameters used to (i)~pretrain the frozen Force-VAE in Stage~1 and
(ii)~fine-tune the $\pi_{0.5}$-based VLA in Stage~2, as well as the precise
configuration of each ablation.

\subsection{Architecture details}
\label{app:impl_arch}

\paragraph{Wrench history processing.}
We use the right-wrist 6-axis wrench stream sampled at $30\,$Hz (down-sampled
from the $100\,$Hz sensor) and apply a causal first-order EMA with
$\alpha = 0.3$. At training, each history is pre-padded with a random-length
Gaussian noise prefix of standard deviation $0.05$ (after quantile
normalization), sampled uniformly with maximum length 1000 frames
($\approx 10\,$s); pre-padding is disabled at inference.

\paragraph{Force-VAE encoder.}
The Force-VAE is a Perceiver-IO
encoder/decoder with $K\!=\!8$ learnable latent tokens. Per-timestep wrench
frames are projected to a $384$-dim token via an input MLP and a Fourier
positional encoding (32 bands, $f_{\max}\!=\!1500$). The encoder stacks
2 cross-attention blocks (1 head, head-dim $64$) feeding 10 latent
self-attention blocks (8 heads, head-dim $32$, dropout
$0.1$); per-latent linear heads emit posterior parameters
$(\mu_k, \log\sigma_k^2) \in \mathbb{R}^{96}$. The decoder mirrors this with
2 cross-attention blocks (8 heads). 

\paragraph{Short state-history projector.}
At each control step we slice the last 10 frames of the joint-state stream at
a stride of 3 dataset frames, i.e.\ offsets
$\{-27, -24, -21, \ldots, -3, 0\}$ relative to the current step
($\approx 0.9\,$s of history at $30\,$Hz). With
$d_s = 16$ (two $7$-DoF arms + two $1$-DoF grippers), the resulting
$10\!\times\!16$ tensor is flattened and projected to a single
$d_h$-dimensional token by a zero-initialized linear layer (no pretraining,
trained end-to-end with the VLA), where $d_h$ is the action-expert hidden width of the $\pi_{0.5}$ base model. 

\paragraph{Memory token injection.}
The frozen Force-VAE encoder produces $Z_f \in \mathbb{R}^{8 \times 96}$;
a zero-initialized linear projector lifts this to $\mathbb{R}^{8 \times d_h}$.
These 8 force-memory tokens are appended to the action-expert suffix
immediately after the 30 noisy-action tokens, followed by the single
short-state token; the noisy-action tokens therefore keep the same RoPE
positions they had during $\pi_{0.5}$ pretraining.

\subsection{Stage~1: Force-VAE pretraining}
\label{app:impl_vae}

We pretrain a single Force-VAE on the right-wrist wrench histories
of all tasks, using inverse-frequency task sampling to prevent the smaller
per-task datasets from being overwhelmed by the larger ones.  
Hyperparameters are summarized in \Cref{tab:vae_hparams}.

\begin{table}[h]
\centering
\caption{\textbf{Force-VAE pretraining hyperparameters} (Stage~1).}
\label{tab:vae_hparams}
\small
\begin{tabular}{ll|ll}
\toprule
hidden / latent width & $384 / 384$ & optimizer & AdamW $(0.9, 0.95)$ \\
per-latent dim $d_z$ & $96$ & grad-clip & $1.0$ \\
\# latent tokens $K$ & $8$ & peak LR & $3\!\times\!10^{-4}$ \\
encoder cross-attn depth & $2$ & warm-up steps & $1\,000$ \\
decoder cross-attn depth & $2$ & total steps & $100\,000$ \\
processor self-attn layers & $10$& batch size & $64\!\times\!8\!=\!512$ \\
Fourier bands & 32 bands & precision & bf16 \\
input dropout & $0.2$ & KL weight $\beta$ & $1\!\times\!10^{-3}$ \\
attention dropout & $0.1$ & free-bits $\lambda$ (per dim) & $0.5$\,nats \\
EMA smoothing $\alpha$ & $0.3$ & noise pre-pad $\sigma$ & $0.05$ \\
noise pre-pad max len & $1000$ &  &  \\
\bottomrule
\end{tabular}
\end{table}

\subsection{Stage~2: VLA fine-tuning}
\label{app:impl_vla}

The Force-VAE encoder is frozen and switched to
eval mode; the projector $W_f$ from the VAE latent to the action-expert
hidden width and the short-state projector are both zero-initialized. All
other components (the VLM, the action expert, and the two projectors) are
trained end-to-end with the flow-matching objective. Per-view image dropout with $p\!=\!0.4$ is applied
independently to the head, left-wrist, and right-wrist RGB streams.

\Cref{tab:vla_hparams} lists the fine-tuning hyperparameters.

\begin{table}[h]
\centering
\caption{\textbf{Stage-2 VLA fine-tuning hyperparameters}, identical across
\name{} and all baselines unless otherwise noted.}
\label{tab:vla_hparams}
\small
\begin{tabular}{ll|ll}
\toprule
\textbf{Model} &  & \textbf{Optimization} &  \\
\midrule
base model & $\pi_{0.5}$ & optimizer & AdamW \\
action chunk horizon $H$ & $30$ & warm-up steps & $1\,000$ \\
action dim & $32$ & peak LR & $5\!\times\!10^{-5}$ \\
flow-matching steps (eval) & $10$ & total steps & $50\,000$ \\
\# force memory tokens $K$ & $8$ & global batch size & $32$ \\
state-window taps & $10$ (stride $3$) & precision & bf16 \\
state window length & $0.9\,$s & hardware & 8 $\times$ A100 (40\,GB) \\
image dropout (per view) & $p\!=\!0.4$ &  &  \\
norm.\ scheme & quantile $q_{01}$/$q_{99}$ &  &  \\
\bottomrule
\end{tabular}
\end{table}

\subsection{Ablation configurations}
\label{app:ablations}

All ablations share the Stage~2 fine-tuning recipe of
\Cref{tab:vla_hparams} (50k steps, same WSD schedule, batch size, image
dropout, data splits, and per-task annotation / norm files) and differ only
in the long-history encoder and which history streams are consumed. 

\paragraph{Force-only (modality ablation).}
The wrench-history branch is unchanged from \name{} (frozen Force-VAE, $K\!=\!8$
tokens, same EMA / pre-pad / quantile-norm pipeline), but the short
state-history projector is removed and the action-expert suffix becomes
\texttt{[noisy actions $\Vert$ 8 force-memory tokens]}. This isolates the
contribution of the long force memory in the absence of any short-horizon
proprioceptive cue.

\paragraph{State-only (modality ablation).}
Identical Perceiver-IO architecture,  trained on long \emph{state} histories of Task~1--3 for
$100\,000$ steps, replaces the Force-VAE. 

\paragraph{GRU long-history encoder (architecture ablation).}
We replace the Perceiver-IO Force-VAE with a single-layer GRU
(\texttt{input\_size}\,=\,6, \texttt{hidden\_size}\,=\,$256$) that
consumes the same EMA-smoothed, pre-padded wrench history and emits its
final hidden state, projected by a single zero-initialized linear layer to
a single token in the action-expert suffix. The short state-history
projector is unchanged. All other hyperparameters are identical to \name{}.
This baseline tests whether a low-capacity recurrent summarizer is
sufficient.

\paragraph{Q-Former long-history encoder (architecture ablation).}
We replace the frozen Force-VAE with a Q-Former cross-attention
module trained end-to-end. The Q-Former has $Q\!=\!8$ learnable query tokens,
$8$ stacked cross-attention layers with $8$ heads and hidden width $256$,
and consumes the same EMA-smoothed, pre-padded long wrench history; its
outputs are projected by a zero-initialized linear layer to $8$ tokens
appended to the action-expert suffix, followed by the same short
state-history token used by \name{}. This matches \name{}'s token budget
($K\!=\!8$) and its position in the suffix; the only changes are
(i) cross-attention instead of a frozen Perceiver-IO VAE and (ii) the
encoder is trained from scratch jointly with the policy rather than
pretrained on wrench reconstruction.

\end{document}